\documentclass{article} 
\pdfoutput=1
\usepackage{iclr2019_conference,times}

\usepackage{amsmath,amsfonts,bm}

\def\eqref#1{equation~\ref{#1}}

\def\1{\bm{1}}

\DeclareMathAlphabet{\mathsfit}{\encodingdefault}{\sfdefault}{m}{sl}
\SetMathAlphabet{\mathsfit}{bold}{\encodingdefault}{\sfdefault}{bx}{n}

\usepackage[utf8]{inputenc}
\usepackage{esvect}
\usepackage{graphicx}
\usepackage{caption}
\usepackage{subcaption}

\usepackage[title]{appendix}
\usepackage{hyperref}
\usepackage[makeroom]{cancel}
\usepackage{amsmath}
\allowdisplaybreaks
\usepackage{dsfont}
\hypersetup{
    colorlinks=true,
    linkcolor=blue,
    filecolor=magenta,      
    urlcolor=cyan,
}
\usepackage{float}
\usepackage{caption}
\usepackage{subcaption}
\usepackage{graphicx}
\usepackage{amssymb}
\usepackage[ruled]{algorithm2e}
\usepackage{caption}

\title{Discovering Hierarchies using Imitation Learning from Hierarchy Aware Policies}

\author{Ameet Deshpande, Harshavardhan P. K., Balaraman Ravindran\\ 
Department of Computer Science and Engineering\\
Indian Institute of Technology Madras\\
\texttt{ameetsd97@gmail.com, harshavardhan864.hk@gmail.com, ravi@cse.iitm.ac.in}\\
}

\iclrfinalcopy 
\begin{document}

\maketitle
\begin{abstract}
    Learning options that allow agents to exhibit temporally higher order behavior has proven to be useful in increasing exploration, reducing sample complexity and for various transfer scenarios. \emph{Deep Discovery of Options} (DDO) is a generative algorithm that learns a hierarchical policy along with options directly from expert trajectories. We perform a qualitative and quantitative analysis of options inferred from DDO in different domains. To this end, we suggest different value metrics like option termination condition, hinge value function error and KL-Divergence based distance metric to compare different methods. Analyzing the termination condition of the options and number of time steps the options were run revealed that the options were terminating prematurely. We suggest modifications which can be incorporated easily and alleviates the problem of \emph{shorter} options and a collapse of options to the same mode.
\end{abstract}

\section{Introduction}
Different frameworks of hierarchical reinforcement learning have been useful in solving complex tasks. Options framework (\cite{dietterich2000hierarchical}) in particular, provides sufficient temporal abstraction to allow actions that have different time scales. They can improve exploration, make learning more sample efficient, aid in transfer or simply help converge to optimal behavior faster especially when the task has sparse reward and long horizon.

The processes of learning options have been widely studied (\cite{barto2003recent}) with some success. Most methods identify salient states (\cite{kulkarni2016hierarchical}) or put constraints on the level of hierarchies. Moreover, learning these options itself takes a lot of data.

\emph{Hierarchical Behavioral Cloning}(HBC) extends the framework of imitation learning to learn options from expert whose trajectories contain information about the option (usually as sub-goal). However, such an expert can't always be available in complex domains where options can't be generated by a human expert.

\emph{Deep Discovery of Options}(DDO) instead views option learning as a probabilistic inference problem where the input is only the flat trajectories of an expert. Using DDO we seek to examine inferred options and apply various metrics to determine their validity and usefulness in learning. We use various similarity metrics to determine how close the generated trajectories are to expert trajectories. We also compare value functions of expert and DDO agent to compare policies directly. We also examine if the options learned to imitate the expert if expert itself uses a hierarchical policy.

We also show simple methods to tackle problems of termination condition being too high for some inferred options. By adding a constant factor to decrease termination probability we can increase the fraction of timesteps taken by options while having similar performance.

Also, by introducing a regularizer to increase KL-divergence between option policies, we can prevent options collapsing to a single mode without reducing the performance.

\section{Related Work}

The problem of \emph{Imitation learning} is to learn a policy based on samples of expert trajectories. Using supervised learning setup to learn action distributions can lead to the problem of compounding errors due to insufficient samples at unsafe or low visitation states. Additional feedback from an expert during training can be incorporated to give more stable policies like in case of DAGGER (\cite{ross2011reduction}).

Hierarchical Reinforcement Learning has proven to be useful paradigm to tackle problems like state, action abstraction , learning low-level skills (\cite{dietterich2000hierarchical,chentanez2005intrinsically}), identifying salient events (\cite{bacon2017option}), exploration, etc (\cite{barto2003recent}). One useful hierarchical framework is to augment options along with primitive actions that perform specific lower level task as described in \cite{sutton1999between,sutton1998intra}.

The problem of learning hierarchical policies from expert feedback has gained popularity recently. Hierarchical Behaviour Cloning by \cite{nejati2006learning} is an extension of passive imitation learning, with options being defined with sub-goals they attempt to reach. The expert needs to augment the primitive actions in trajectories with options. The work by \cite{le2018hierarchical} extends DAGGER to hierarchical setup where expert gives additional feedback such as suggesting sub-goals as options after inspecting trajectories.

Deep Discovery of Options by \cite{fox2017multi} and another work by \cite{smith2018inference} solve the problem of option discovery using probabilistic inference approach assuming a Hidden-Markov model and using EM algorithms (\cite{welch2003hidden}) to infer option policies. A follow-up work (\cite{krishnan2017ddco}) extends this framework to continuous actions tasks.

\section{Preliminaries}

We use the usual notations for Markov Decision Process (MDP) which is described by $<S,A,R,T,p_0,\gamma>$ where $S$ is set of states, $A$ the set of actions, $R(s,a)$ the reward function on taking an action $a$ at given state $s$ and transition function $T(s,a)=P(s'|s,a)$. $\gamma$ is the discount factor for rewards.

$\pi(a|s)$ gives the next action distribution. The state-action value function $Q(s,a)$ is the expected total return
\[Q(s,a)=\mathbf{E}[R(s,a)+\gamma\sum_{s'\in S}P(s'|s,a)\underset{a'}{\max}Q(s',a')]\]
Thus, optimal policy is $\pi(a|s)\underset{a}{\arg\max}Q(s,a)$.

In DQN (\cite{mnih2013playing}), a neural network approximates $Q(s,a)$ and uses the above bellman equation to update function for every transition.

\subsection{Hierarchical Behavioural Cloning}
In usual behavioral cloning the task is to learn policy from expert trajectories $\{s_0,a_0,s_1,a_1,\dots,s_T\}$. Hierarchical Behavioural Cloning (\cite{nejati2006learning}) extends this to options setup where options are mapped to sub-goals  and at each state $s_t$ expert is assumed to select an option $g_t$ and from then on follow trajectory $\tau_t$ according to option policy upto termination of option. Thus the trajectory of expert looks like $\{(s_0,g_0,\tau_0),(s_1,g_1,\tau_1)\dots,s_T\}$ and agent learn a hierarchical policy to maximize likelihood of expert trajectories. 

\begin{figure}[H]
    \centering
    \includegraphics[width=.6\textwidth]{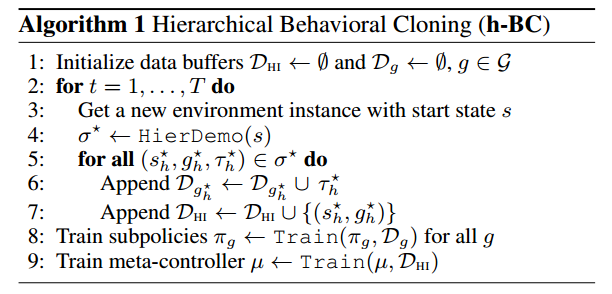}
    
\end{figure}

\subsection{Deep Discovery of Options}
Deep Discovery of Options (DDO) uses the behavioral cloning framework to automatically infer useful polices from flat trajectories of expert. Using the flat trajectories $\xi=(s_0,a_0,s_1,a_1,\dots,s_T)$ as observable variables they assume a hidden markov model with latent variables $h_t$ and $b_t$ being current option being executed and termination probability of current state $s_t$ under option $h_t$.

Then, the latent variables are $\zeta=(b_0,h_0,b1,h_1,\dots,b_{T-1},h_{T-1})$. Using a EM-Algorithm similar to Baum-Welch DDO computes options policy $\pi_h(.|s)$, termination probabilities for options $\phi_h(s)$ and meta policy $\eta(.|s)$ parameterized by $\theta$.

It uses the Expectation gradient trick.

\[L(\theta;\xi)=\log P(\xi|\theta) = \log P_{\theta}(\xi)\]
\[{\begin{split}
    \nabla_{\theta}L(\theta;\xi) & =\nabla_{\theta}\log P_{\theta}(\xi) = \frac{\nabla_{\theta}P_{\theta}(\xi)}{P_{\theta}(\xi)}\\
    & = \frac{1}{P_{\theta}(\xi)}\sum_{\zeta}\nabla_{\theta}P_{\theta}(\xi,\zeta)\\
    & = \frac{1}{P_{\theta}(\xi)}\sum_{\zeta}P_{\theta}(\xi,\zeta)\nabla_{\theta}\log P_{\theta}(\xi,\zeta) = E_{\xi|\zeta}[\nabla_{\theta}\log P_{\theta}(\zeta,\xi)]\\
    & =\nabla_{\theta}\log \eta(h_0|s_0) + \sum_{t=0}^{T-1}\nabla_{\theta}\log \pi_{h_t}(a_t|s_t)
    +\sum_{t=1}^{T}\nabla_{\theta}\log P_{\theta}(b_t,h_t|h_{t-1},s_t)
    \end{split}
    }\]

Then a forward-backward algorithm is used, reminiscent of Baum-Welch:
{\[ u_{t}(h) = P_{\theta}(h_t=h|\xi)  \]}
  {\[ v_{t}(h) = P_{\theta}(b_t=1, h_t=h|\xi)  \]}
    {\[ w_{t}(h) = P_{\theta}(b_{t+1}=0,h_t=h|\xi)  \]}
    { \[{\begin{split}\nabla_{\theta}L(\theta;\xi) & = \sum_{h\in H} (\sum_{t=0}^{T-1}( v_{t}(h) \nabla_{\theta} \log \eta(h|s_t) + u_t(h) \nabla_{\theta} \log \pi_h(a_t|s_t) )\\ & + (\sum_{t=0}^{T-2}( (u_t(h)-w_t(h)) \nabla_{\theta} \log \psi_h(s_{t+1}) + w_t(h)\nabla_{\theta} \log (1-\psi_h(s_{t+1})) )   ))  \end{split}}\]}
The gradient computed above can then be used in
any stochastic gradient descent algorithm.

$u,v,w$ can be written in terms of the forward and backward probability
    {\[\phi_t(h) = P_{\theta}(s_0,a_0,\dots,s_t,h_t=h) \]}
    {\[ \phi_{t+1}(h) = \sum_{h'\in H} \phi_t(h')\times term_1\times \psi_h'(s_{t+1}) + \phi_t(h)\times term_2\times  (1-\psi_h(s_{t+1})) \]}
    {\[\omega_t(h) = P_{\theta}(a_t,s_{t+1},\dots,s_T|s_t,h_t=h) \]}
    {\[ u_t(h) = \dfrac{1}{P(\xi)}\phi_t(h)\omega_t(h) \]}
    {
    The dynamics term, $p_o(s_0) \prod_{t=0}^T p(s_{t+1}|s_t,a_t)$ gets cancelled in these normalizations, which allows us to ignore the dynamics}

\subsection{Hierarchical Deep Q-Network}
Hierarchical Deep Q-Network (h-DQN) (\cite{kulkarni2016hierarchical}) is an extension of DQN (\cite{mnih2013playing}) to learn option policy and meta-policy simultaneously. The options are defined, as in HBC, with respect to sub-goal states and intrinsic reward is used to learn option specific value functions $Q(s,a;g)$. The meta policy is learned via action values w.r.t options $Q(s,g)$. Each of the two value functions are approximated using separate networks. The options network is updated using normal TD-error
\[L_{1} = \mathbf{E}_{(s,a,r,s',g)\sim \mathcal{D_1}}[(r+\gamma\underset{a'}{Q(s',a',g;\theta')}-Q(s,a,g;\theta))^2]\]
where $r$ is intrinsic reward, while meta policy network uses SMDP Q-learning (\cite{dietterich2000hierarchical}) updates:
\[L_{2} = \mathbf{E}_{(s,g,f,s',t')\sim \mathcal{D_2}}[(f+\gamma^{t'}\underset{g'}{Q(s',g';\phi')}-Q(s,g;\phi))^2]\]

where $f$ is the discounted reward during execution of option from the environment (only extrinsic reward).

\section{Learning options in Grid-worlds}

We first train DDO agent in grid world and analyze the options inferred.

\subsection{Method}
The method for training is summarized in algorithm \ref{alg:grid1}.
Note that meta-policy uses both options and primitive actions.

\begin{algorithm}[h]

\SetAlgoLined
\SetKwInOut{Input}{input}
\SetKwInOut{Output}{output}

\For{$i\leftarrow[T]$}{
Initialize a random start and goal state\;
Learn $Q_{expert}$ using value iteration.\;
Sample trajectories using $Q_{expert}$ to $\mathcal{D}$\;
}

Use trajectories in $D$ to infer options\;
Use SMDP Q-learning to learn meta-policy from options and primitive actions\;
\caption{Training in Grid world tasks}
\label{alg:grid1}
\end{algorithm}

\subsection{Analysis of Inferred Options}
We used 4 to 6 options depending on the size of the grid world. In case the number of rooms is more than 2, we used 6 options, as in the case of the four-room grid.
Some of the inferred options are visualized in Appendix \ref{app:infer1}. We can see quite a bit of variation in option policies. However, the termination probability was seen to be very high as seen in Table \ref{tab:4room}. Mostly, options were not executed for more than one or two timesteps.

\begin{table}[h]
    \centering
    \begin{tabular}{|c|c|c|}
    \hline
        \textbf{Option} & \textbf{Mean $\beta$} & \textbf{Variance}\\
        \hline
        1 &  0.64 & 0.020\\
        2 &  0.69 & 0.011\\
        3 &  0.81 & 0.006\\
        4 &  0.70 & 0.010\\
        5 &  0.24 & 0.009\\
        6 &  0.27 & 0.009\\
         \hline
    \end{tabular}
    \caption{Option lengths in 4-room world}
    \label{tab:4room}
\end{table}   

\begin{figure}[h]
    \centering
    \begin{subfigure}{.5\textwidth}
      \centering
      \includegraphics[width=\linewidth]{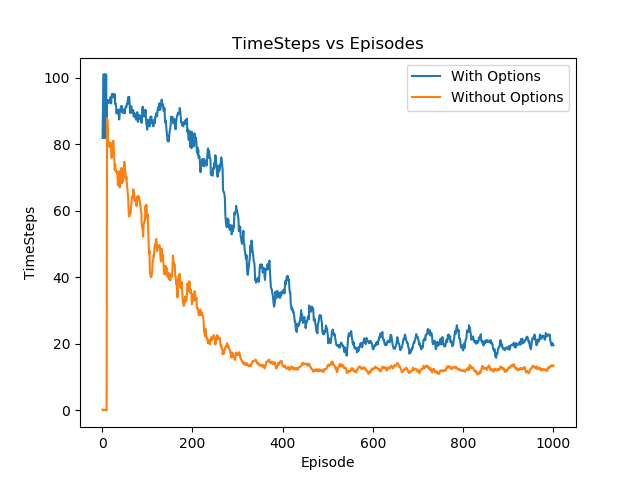}
    \end{subfigure}%
    
    \caption{Learning curve on 4-room environment}
    \label{fig:learn}
    \end{figure}

We also noticed that Q-learner learned faster without using options (see Figure \ref{fig:learn}). This might indicate that options were not necessarily used for better exploration. Instead, options were similar to primitive actions in behavior due to high termination condition and hence updating Q values for options along with primitives might have increased time to reach optimal behavior.

\paragraph{Similarity of trajectories}
In order to assess how similar trajectories of agent and expert are we used the KL-Divergence over action probabilities as a metric for similarity. The results are shown in Table \ref{tab:kl1}.

\begin{table}[h]
    \centering
    \begin{tabular}{|c|c|}
    \hline
        \textbf{Environment} & \textbf{CE-error} \\
        \hline
        Roundabout & 0.270\\
        4room & 0.331\\
        Hallway & 1.040\\
        experiment2 & 0.276\\
         \hline
    \end{tabular}
    \caption{KL Divergence between policies of expert and DDO agent}
    \label{tab:kl1}
\end{table}

A random policy for this task would give a KL-Divergence value of 0.69. In many environments, though the CE-error is high, it is doing better than a random policy. But the error shows that the trajectories may not be very similar.


\paragraph{Hinge loss of value function}
Next, we see if the value function learned by the expert in given task matches that of the agent. We use hinge loss of agent's value function w.r.t that of the expert. Hinge loss measures the difference if the value function of the agent is lesser than that of expert policy at a state. Thus, it measures the maximal marginal difference.
\[L_{hinge}=\frac{1}{|\mathcal{S}|}\sum_{s\in\mathcal{S}}(\min (V(s),V^*(s)) - V^*(s))^2\]

Lower the loss, closer is value function of the agent to that of expert (which is learned via value iteration). 


The results are summarized in Table \ref{tab:hinge}.

\begin{table}[h]
    \centering
    \begin{tabular}{|c|c|}
    \hline
        \textbf{Environment} & \textbf{Hinge error} \\
        \hline
        Roundabout & 0.364\\
        4room & 0.468\\
        Hallway & 0.198\\
        experiment2 & 0.372\\
         \hline
    \end{tabular}
    \caption{Hinge loss between value functions}
    \label{tab:hinge}
\end{table}

\subsection{Learning from a hierarchical expert}
\begin{figure}[h]
    \centering
    \begin{subfigure}{.5\textwidth}
      \centering
      \includegraphics[width=\linewidth]{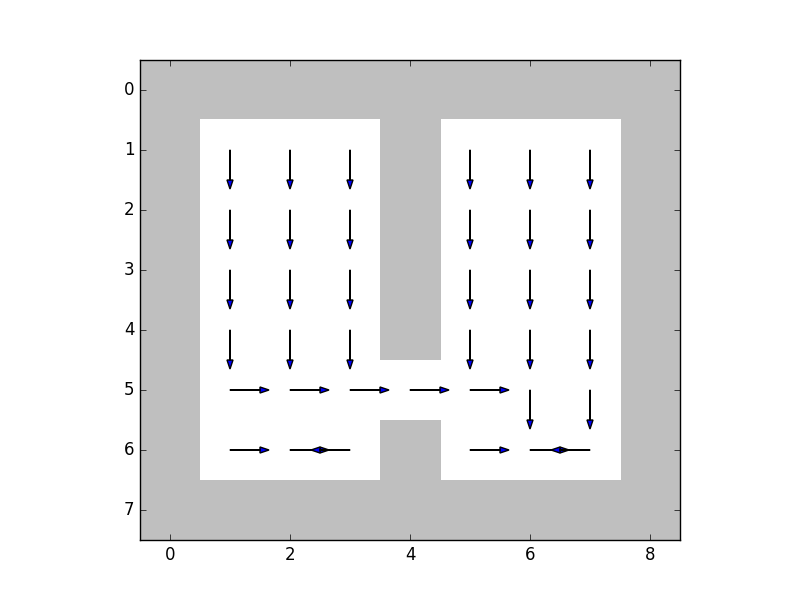}
      \caption{Expert option}
    \end{subfigure}%
    \begin{subfigure}{.5\textwidth}
      \centering
      \includegraphics[width=\linewidth]{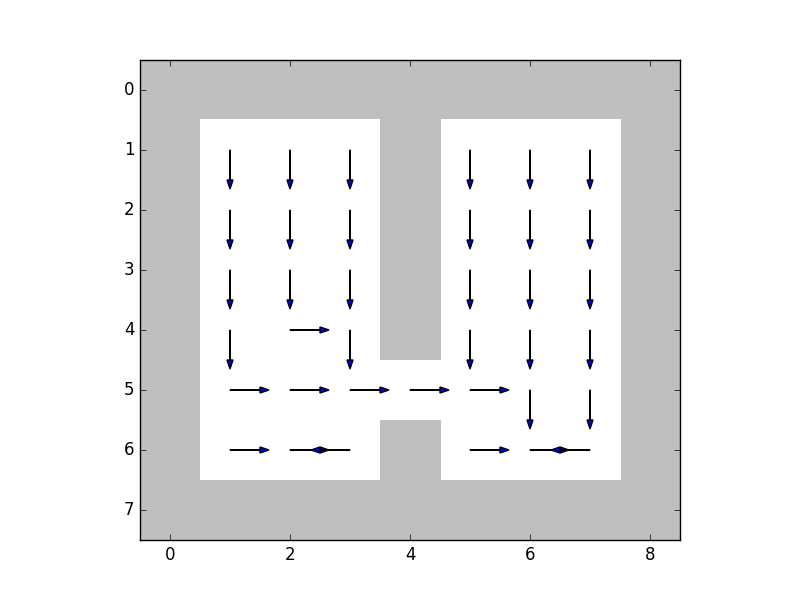}
      \caption{Inferred option}
    \end{subfigure}
    \begin{subfigure}{.5\textwidth}
      \centering
      \includegraphics[width=\linewidth]{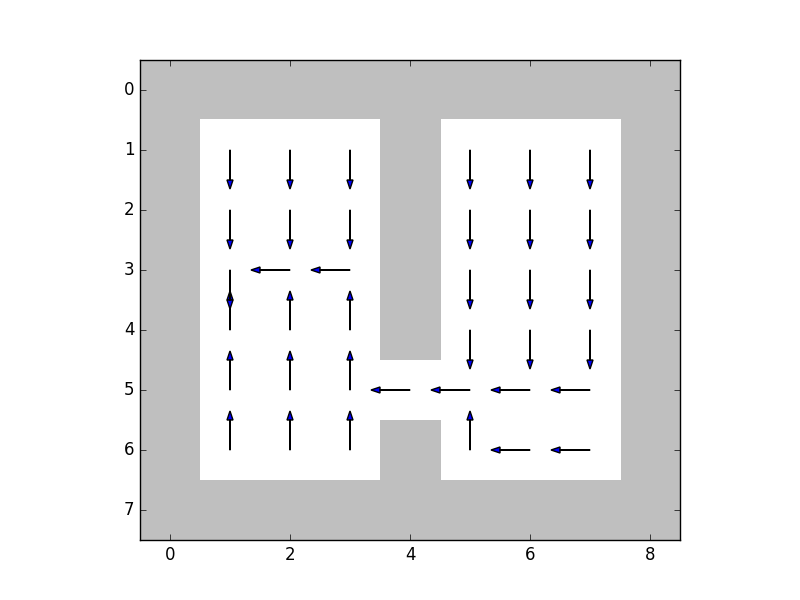}
      \caption{Expert option}
    \end{subfigure}%
    \begin{subfigure}{.5\textwidth}
      \centering
      \includegraphics[width=\linewidth]{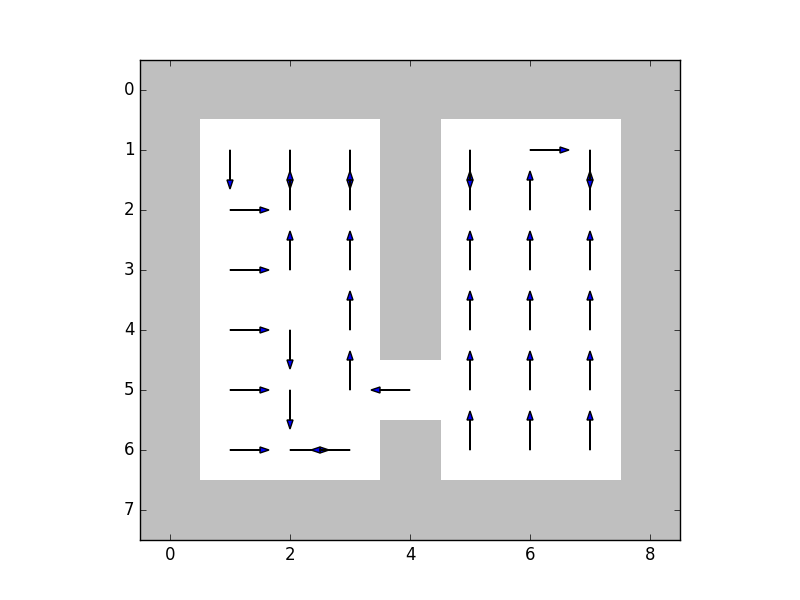}
      \caption{Inferred option}
    \end{subfigure}
    \caption{Comparing expert option and inferred option}
    \label{fig:DDOexpert}
    \end{figure}

So far the expert was trained using value iteration and didn't exhibit hierarchical behavior. We now train using hierarchical expert by using previously trained DDO agent as an expert. We notice that there is not much similarity between expert and inferred options as seen in figure \ref{fig:DDOexpert}.

\begin{figure}[h]
    \centering
    \begin{subfigure}{.5\textwidth}
      \centering
      \includegraphics[width=\linewidth]{interim2/Images/experiment10.png}
      \caption{Expert}
    \end{subfigure}%
    \begin{subfigure}{.5\textwidth}
      \centering
      \includegraphics[width=\linewidth]{interim2/Images/experiment13_trajs.png}
      \caption{Inferred}
    \end{subfigure}
    \begin{subfigure}{.5\textwidth}
  \centering
  \includegraphics[width=\linewidth]{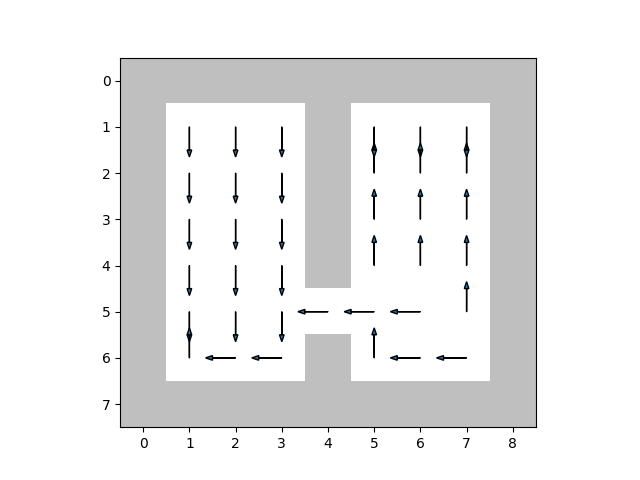}
  \caption{Expert}
\end{subfigure}%
\begin{subfigure}{.5\textwidth}
  \centering
  \includegraphics[width=\linewidth]{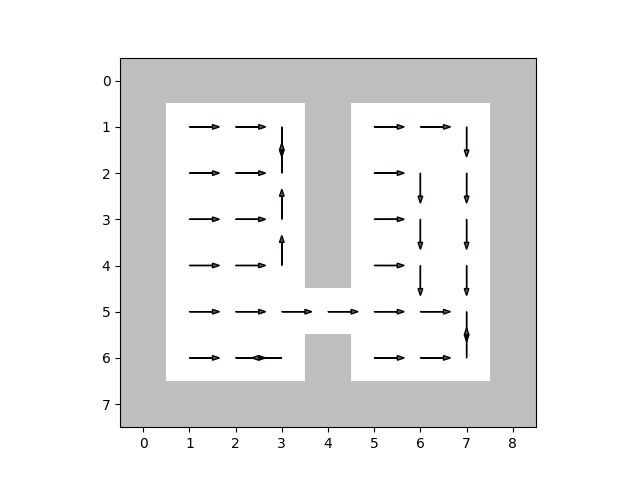}
  \caption{Inferred}
\end{subfigure}
\caption{Comparing expert option and inferred option in case of hand-coded options for expert}
    \label{fig:DDOhand}
    \end{figure}

We also tried to hand-code options to expert policy and first train it using SMDP Q-learning and then training DDO agent using trajectories of the expert. Again we didn't notice many similarities as seen in Figure \ref{fig:DDOhand}.

However, we noticed that the hinge loss decreased to 0.014 and KL-Divergence between policies was found to be 0.19. Hence, DDO agent better emulated the hierarchical expert.


\subsection{Overcoming large termination probability}
A major obstacle to using inferred options for longer time-steps in DDO agent was that termination probabilities were simply too high. Hence, we added a multiplicative factor $\alpha\in (0,1]$ to inferred termination probability to decrease it before learning the meta-policy. This allows options to execute for a larger number of timesteps before termination.
\[\beta_{new} = \beta\times \alpha\]



As expected, the fraction of time options was used increased with decreasing $\alpha$. However, the hinge loss also slightly increased (see Table \ref{tab:alpha}).


\begin{table}[H]
    \centering
    \begin{tabular}{|c|c|c|c|c|}
        \hline
        $\boldsymbol{\alpha}$ & \textbf{Median $\beta_{new}$} & \textbf{Mean $\beta_{new}$} & \textbf{$\%$ of Option Time} & \textbf{Hinge loss}\\ \hline
        0.1 & 5.3 & 6.4 & 0.66 & 0.41\\ \hline
        0.2 & 4 & 4 & 0.58 & 0.37\\ \hline
        0.4 & 2.98 & 5.57 & 0.68 & 0.40\\ \hline
        0.7 & 1.4 & 1.76 & 0.60 & 0.37\\ \hline
        1.0 & 0.5 & 4.34 & 0.42 & 0.369\\ \hline
    \end{tabular}
    \caption{Comparing effect of $\alpha$ on termination probability and learned value function}
    \label{tab:alpha}
\end{table}

\subsection{Avoiding mode collapse of options}
We noticed that multiple options inferred are very similar. Hence, to increase the diversity of options we added a regularizer to increase the KL-Divergence between policies of options.



Hence, our new loss function becomes
\[-L(\theta;\xi) - \lambda\times E_{\rho}[\sum_{i=1}^{k} \sum_{j=1, j\neq i}^{k} KL\_Divergence(option_i, option_j)]\]

where $L(\theta;\xi)$ is log likelihood of DDO parameters $\theta$ given expert trajectories $\xi$ and $\lambda$ determines importance of KL-divergence term. The effect of $\lambda$ on inferred policy is summarized in Table \ref{tab:lamda}.

\begin{table}[H]
    \centering
    \begin{tabular}{|c|c|c|}
    \hline
        \textbf{$\boldsymbol{\lambda}$} & \textbf{Median steps for options} & \textbf{Error} \\ \hline
        0.001 & 1.18 & 0.368\\ \hline
        0.01 & 1.0 & 0.378\\ \hline
        0.1 & 4.1 & 0.42\\ \hline
        0.2 & 4.5 & 0.40\\ \hline
        \textbf{0.3} & \textbf{5.0} & \textbf{0.41}\\ \hline
        0.4 & 4.8 & 0.40\\ \hline
        0.5 & 6.0 & 0.44\\ \hline
        0.6 & 2.85 & 0.37\\ \hline
        0.7 & 1.0 & 0.37\\ \hline
    \end{tabular}
    \caption{Effect of KL-divergence term on DDO policy}
    \label{tab:lamda}
\end{table}

We see that setting $\lambda$ too high or low makes the median steps per option low. With $\lambda$ too high, options learned may not be useful enough and only primitive actions may be chosen. With $\lambda$ too low, no attention is paid to the KL-Divergence term and the options learned might all be similar because of Mode Collapse.


\section{Learning options in Atari Domain}
We also used DDO on PONG and KRULL games to analyze the options inferred in domains with large state space. We chose trained A3C as expert policy.

\subsection{Training}
We used HDQN framework to train the meta-policy after learning options. Unlike in actual HDQN setup described in \cite{kulkarni2016hierarchical} we don't train the network to learn option policies also, but only train the network for meta-controller to learn policy over options and primitive actions as described in Algorithm \ref{alg:HDQN}.

\begin{algorithm}[h]

\SetAlgoLined
\SetKwInOut{Input}{input}
\SetKwInOut{Output}{output}
\Input{Network parameters $\theta$, Option policies $opt=\{\pi_{h1},\pi_{h2},\dots,\pi_{hO}\}$, Termination proabilities $term=\{\phi_{h1},\phi_{h1},\dots,\phi_{hO}\}$, other DQN hyperparameters}
Initialize target and current network with weights $\theta'=\theta$\;
\For{$ep\leftarrow[N]$}{
    $s_t\leftarrow$ Reset environment\;
    \For{$t\leftarrow[T]$}{
        Select $h_t$ using $\epsilon-$greedy on $Q^{\theta'}(s_t,.)$\;
        \If{$h_t$ is option}
        {$done\leftarrow False$\
        $t',R\leftarrow 0$\;
            \While{not $done$}{
                $a_{t'}\leftarrow \pi_{h_t}(.|s_{t+t'})$\;
                Get $s_{t+t'}, r_{t+t'}$\;
                $R+=R+\gamma^{t'}r_{t+t'}$\;
                Sample $b\leftarrow [0,1]$\;
                \If{$b<\phi_{h_t}(s_{t+t'})$ or episode terminates or exceeds $max_steps$}{
                    $done\leftarrow True$
                }
            }
        
        }
        \Else{
            $t'\leftarrow 1$\;
            Get $s_{t+1},R$ from environment\;
        }
        $D\leftarrow D\cup \{s_t,h_t,R,s_{t+t'},t'\}$\;
        Sample $k$ sampled from $D$\;
        Batch update DQN using squared TD error loss $(R+\gamma^{t'}\max_hQ^{\theta}(s_{t+t'},a)-Q^{\theta*}(s_t,h_t))^2$\;
    }
}

\caption{TrainHDQN}
\label{alg:HDQN}
\end{algorithm}

Using the simple training procedure as in case of Grid-world domains where DDO options are first learned and then meta-policy is learned did not converge quickly in our case. So we used an iterated training procedure as described in Algorithm \ref{alg:DDOHDQN}. First, we sampled $T=1000$ trajectories from trajectories buffer. In each of the $N=10$ iterations, we first refined DDO parameters from sampled trajectories and then refined meta-policy of HDQN. We added $T'=100$ samples from agent to trajectory buffer.

\begin{algorithm}[h]

\SetAlgoLined
\SetKwInOut{Input}{input}
\SetKwInOut{Output}{output}
\Input{Number of options:$O$,global steps:$N$,Sample size $T$, Sample ddo trajectores $T'$ other hyperparameters}
Learn expert policy $\pi_{exp}$ using A3C\;
Initialize DQN $Q^{\theta}(s,a),a\in actions\cup options$\;
Sample $T$ trajectories $D=\{t_1,t_2,\dots,t_{T}\}$\;
\For{$i\leftarrow N$}{
    Sample $T$ trajectories form $D$\;
    Train DDO parameters for option policies $opt=\{\pi_{h1},\pi_{h2},\dots,\pi_{hO}\}$ and termination conditions $term=\{\phi_{h1},\phi_{h1},\dots,\phi_{hO}\}$ using sampled trajectories\;
    $\theta\leftarrow TrainHDQN(\theta,opt,term)$\;
    Rollout $T'$ trajectories from HDQN $D'=\{t'_1,t'_2,\dots,t'_{T}\}$\;
    $D\leftarrow D\cup D'$ 
}

\caption{DDO + HDQN}
\label{alg:DDOHDQN}
\end{algorithm}

\subsection{Analysis of inferred options}
We used 10 options for PONG and 20 options for KRULL. 

Qualitative analysis found only up to 3 options in each game that looked useful. Rest of the options usually involved high-frequency periodic oscillations in state transitions.

\begin{figure}[h]
    \centering
    \begin{subfigure}{0.8\textwidth}
                \includegraphics[width=.9\textwidth]{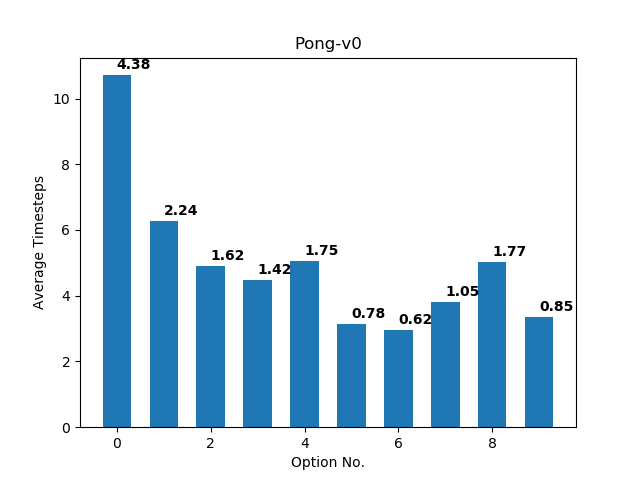}
                \caption{Pong options}
                \label{fig:gull}
        \end{subfigure}
    \begin{subfigure}{0.8\textwidth}
                \includegraphics[width=.9\textwidth]{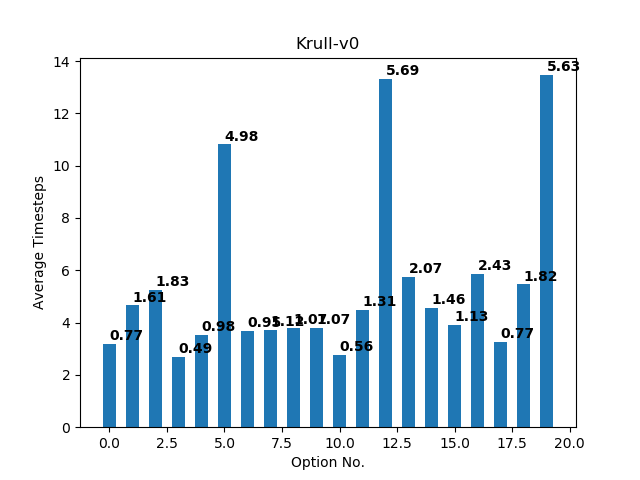}
                \caption{Krull options}
                \label{fig:gull}
        \end{subfigure}
    
    \caption{Analysis of options in PONG and KRULL\\ The values at top of bars depict the standard deviation.}
    \label{fig:opt}
\end{figure}

We provide the average time-steps for each option in trained DDO agent along with standard deviation in figure \ref{fig:opt}.

\section{Discussion}
We have analyzed options inferred by DDO algorithm in some grid-world and Atari domains. We used metrics like KL-divergence between policies, hinge loss between value functions to assess similarity in behavior of agent and expert.

We found that termination probabilities inferred are too high.

We used ad-hoc methods to alleviate problems like high termination probabilities for options and mode collapse of options. These solutions seem to make the policies learned use options inferred more frequently and for a longer fraction of trajectories. We multiplied with a constant factor to decrease termination probability for states. We introduced KL-divergence as a regularizer to increase the variety of options and prevent mode collapse.

In case of training DDO agent in Atari domain, alternating between training DDO parameters for options and training HDQN for meta-policy for several time-steps helped to learn faster than just learning meta-policy at end of DDO option inference.

Importance of options can also be validated by salient states or events discovered, usefulness in transfer to slightly different tasks in the same environment, state spaces where they operate, etc. 
Creating metrics for continuous actions domains could also be a good next step.

\newpage

\bibliographystyle{iclr2019_conference}
\bibliography{references}

\begin{thebibliography}{15}
\providecommand{\natexlab}[1]{#1}
\providecommand{\url}[1]{\texttt{#1}}
\expandafter\ifx\csname urlstyle\endcsname\relax
  \providecommand{\doi}[1]{doi: #1}\else
  \providecommand{\doi}{doi: \begingroup \urlstyle{rm}\Url}\fi

\bibitem[Bacon et~al.(2017)Bacon, Harb, and Precup]{bacon2017option}
Pierre-Luc Bacon, Jean Harb, and Doina Precup.
\newblock The option-critic architecture.
\newblock In \emph{AAAI}, pp.\  1726--1734, 2017.

\bibitem[Barto \& Mahadevan(2003)Barto and Mahadevan]{barto2003recent}
Andrew~G Barto and Sridhar Mahadevan.
\newblock Recent advances in hierarchical reinforcement learning.
\newblock \emph{Discrete event dynamic systems}, 13\penalty0 (1-2):\penalty0
  41--77, 2003.

\bibitem[Chentanez et~al.(2005)Chentanez, Barto, and
  Singh]{chentanez2005intrinsically}
Nuttapong Chentanez, Andrew~G Barto, and Satinder~P Singh.
\newblock Intrinsically motivated reinforcement learning.
\newblock In \emph{Advances in neural information processing systems}, pp.\
  1281--1288, 2005.

\bibitem[Dietterich(2000)]{dietterich2000hierarchical}
Thomas~G Dietterich.
\newblock Hierarchical reinforcement learning with the maxq value function
  decomposition.
\newblock \emph{Journal of Artificial Intelligence Research}, 13:\penalty0
  227--303, 2000.

\bibitem[Fox et~al.(2017)Fox, Krishnan, Stoica, and Goldberg]{fox2017multi}
Roy Fox, Sanjay Krishnan, Ion Stoica, and Ken Goldberg.
\newblock Multi-level discovery of deep options.
\newblock \emph{arXiv preprint arXiv:1703.08294}, 2017.

\bibitem[Krishnan et~al.(2017)Krishnan, Fox, Stoica, and
  Goldberg]{krishnan2017ddco}
Sanjay Krishnan, Roy Fox, Ion Stoica, and Ken Goldberg.
\newblock Ddco: Discovery of deep continuous options for robot learning from
  demonstrations.
\newblock In \emph{Conference on Robot Learning}, pp.\  418--437, 2017.

\bibitem[Kulkarni et~al.(2016)Kulkarni, Narasimhan, Saeedi, and
  Tenenbaum]{kulkarni2016hierarchical}
Tejas~D Kulkarni, Karthik Narasimhan, Ardavan Saeedi, and Josh Tenenbaum.
\newblock Hierarchical deep reinforcement learning: Integrating temporal
  abstraction and intrinsic motivation.
\newblock In \emph{Advances in neural information processing systems}, pp.\
  3675--3683, 2016.

\bibitem[Le et~al.(2018)Le, Jiang, Agarwal, Dud{\'\i}k, Yue, and
  Daum{\'e}~III]{le2018hierarchical}
Hoang~M Le, Nan Jiang, Alekh Agarwal, Miroslav Dud{\'\i}k, Yisong Yue, and Hal
  Daum{\'e}~III.
\newblock Hierarchical imitation and reinforcement learning.
\newblock \emph{arXiv preprint arXiv:1803.00590}, 2018.

\bibitem[Mnih et~al.(2013)Mnih, Kavukcuoglu, Silver, Graves, Antonoglou,
  Wierstra, and Riedmiller]{mnih2013playing}
Volodymyr Mnih, Koray Kavukcuoglu, David Silver, Alex Graves, Ioannis
  Antonoglou, Daan Wierstra, and Martin Riedmiller.
\newblock Playing atari with deep reinforcement learning.
\newblock \emph{arXiv preprint arXiv:1312.5602}, 2013.

\bibitem[Nejati et~al.(2006)Nejati, Langley, and Konik]{nejati2006learning}
Negin Nejati, Pat Langley, and Tolga Konik.
\newblock Learning hierarchical task networks by observation.
\newblock In \emph{Proceedings of the 23rd international conference on Machine
  learning}, pp.\  665--672. ACM, 2006.

\bibitem[Ross et~al.(2011)Ross, Gordon, and Bagnell]{ross2011reduction}
St{\'e}phane Ross, Geoffrey Gordon, and Drew Bagnell.
\newblock A reduction of imitation learning and structured prediction to
  no-regret online learning.
\newblock In \emph{Proceedings of the fourteenth international conference on
  artificial intelligence and statistics}, pp.\  627--635, 2011.

\bibitem[Smith et~al.(2018)Smith, Hoof, and Pineau]{smith2018inference}
Matthew Smith, Herke Hoof, and Joelle Pineau.
\newblock An inference-based policy gradient method for learning options.
\newblock In \emph{International Conference on Machine Learning}, pp.\
  4710--4719, 2018.

\bibitem[Sutton et~al.(1998)Sutton, Precup, and Singh]{sutton1998intra}
Richard~S Sutton, Doina Precup, and Satinder~P Singh.
\newblock Intra-option learning about temporally abstract actions.
\newblock In \emph{ICML}, volume~98, pp.\  556--564, 1998.

\bibitem[Sutton et~al.(1999)Sutton, Precup, and Singh]{sutton1999between}
Richard~S Sutton, Doina Precup, and Satinder Singh.
\newblock Between mdps and semi-mdps: A framework for temporal abstraction in
  reinforcement learning.
\newblock \emph{Artificial intelligence}, 112\penalty0 (1-2):\penalty0
  181--211, 1999.

\bibitem[Welch(2003)]{welch2003hidden}
Lloyd~R Welch.
\newblock Hidden markov models and the baum-welch algorithm.
\newblock \emph{IEEE Information Theory Society Newsletter}, 53\penalty0
  (4):\penalty0 10--13, 2003.

\end{thebibliography}

\newpage
\begin{appendices}

\section{Visualization of inferred policies}\label{app:infer1}

We show some of the inferred policies in grid-world tasks using expert trained via value iteration.

\begin{figure}[H]
    \centering
    \begin{subfigure}{.5\textwidth}
      \centering
      \includegraphics[width=\linewidth]{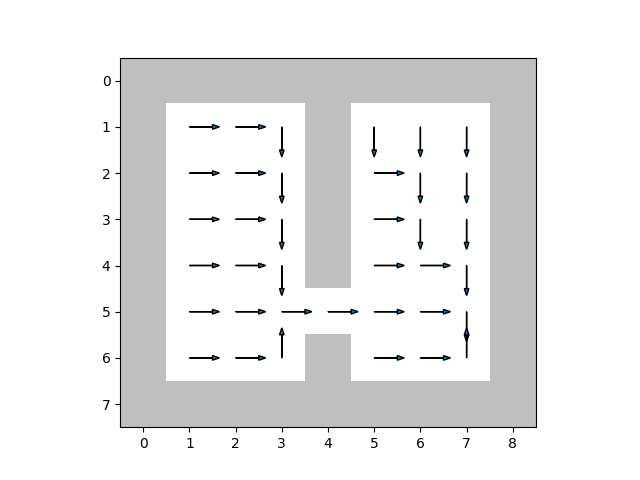}
    \end{subfigure}%
    \begin{subfigure}{.5\textwidth}
      \centering
      \includegraphics[width=\linewidth]{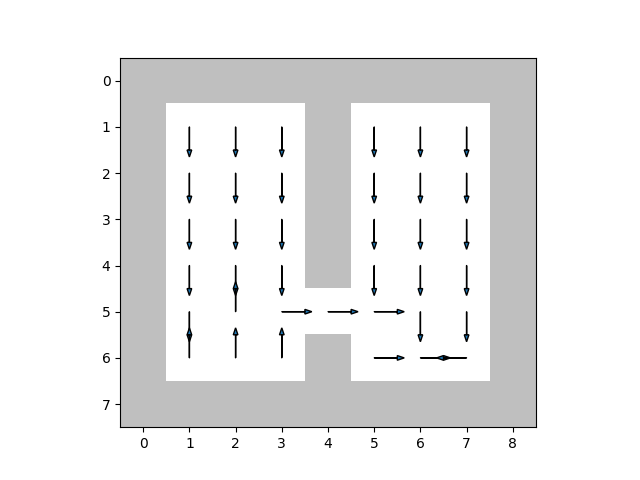}
    \end{subfigure}
    \end{figure}    
    
   \begin{figure}[H]
    \centering
    \begin{subfigure}{.5\textwidth}
      \centering
      \includegraphics[width=\linewidth]{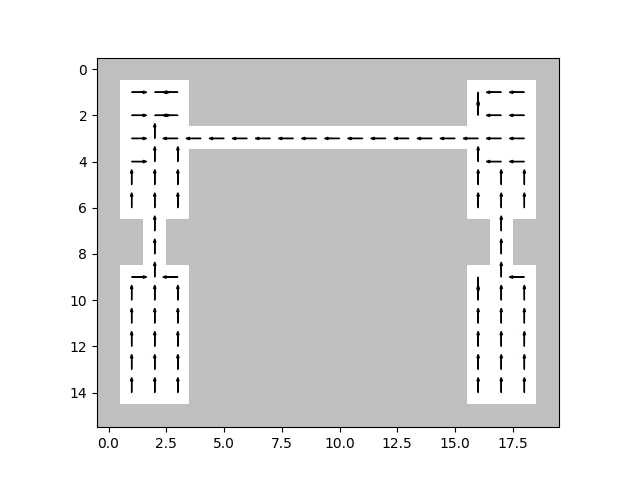}
    \end{subfigure}%
    \begin{subfigure}{.5\textwidth}
      \centering
      \includegraphics[width=\linewidth]{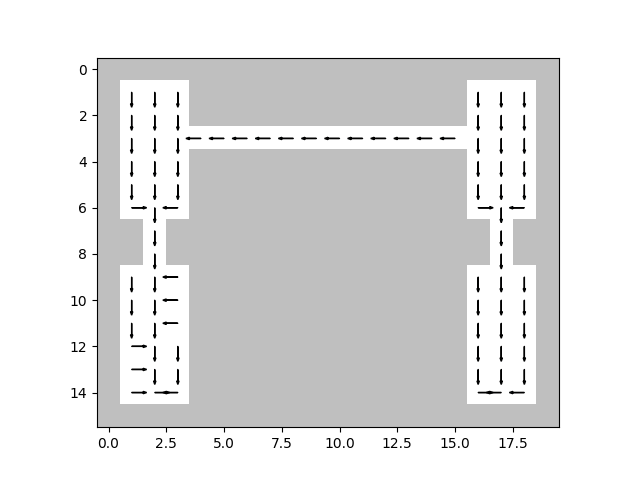}
    \end{subfigure}
    \end{figure}

    \begin{figure}[H]
    \centering
    \begin{subfigure}{.5\textwidth}
      \centering
      \includegraphics[width=\linewidth]{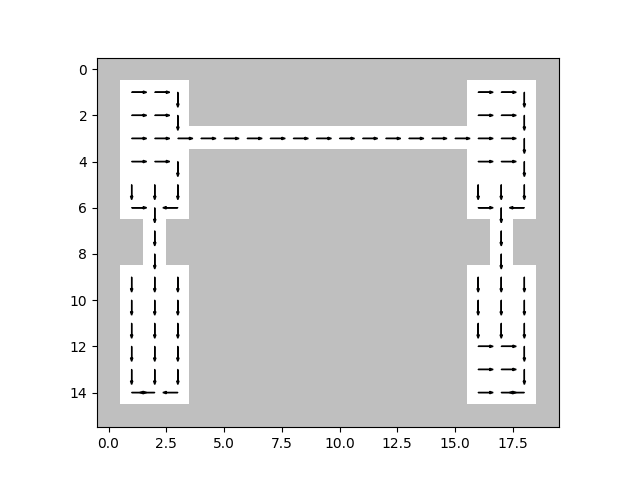}
    \end{subfigure}%
    \begin{subfigure}{.5\textwidth}
      \centering
      \includegraphics[width=\linewidth]{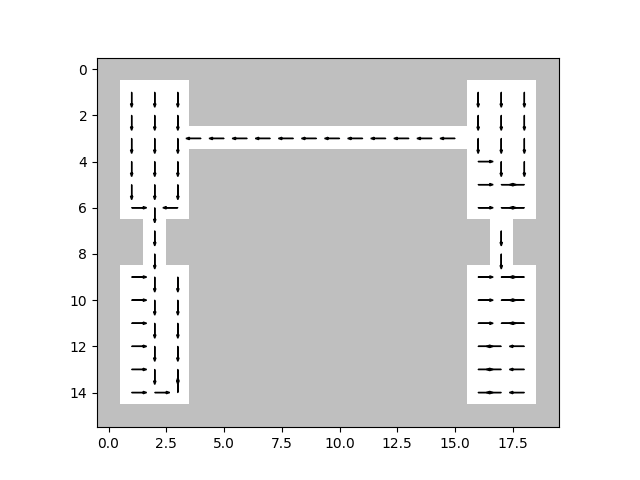}
    \end{subfigure}
    \end{figure}

    \begin{figure}[H]
    \centering
    \begin{subfigure}{.5\textwidth}
      \centering
      \includegraphics[width=\linewidth]{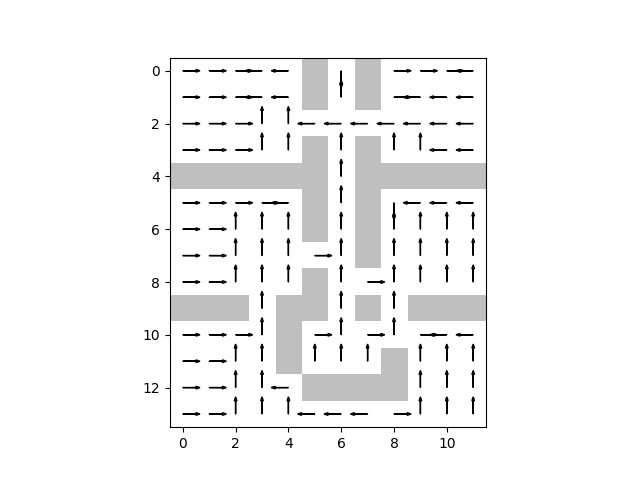}
    \end{subfigure}%
    \begin{subfigure}{.5\textwidth}
      \centering
      \includegraphics[width=\linewidth]{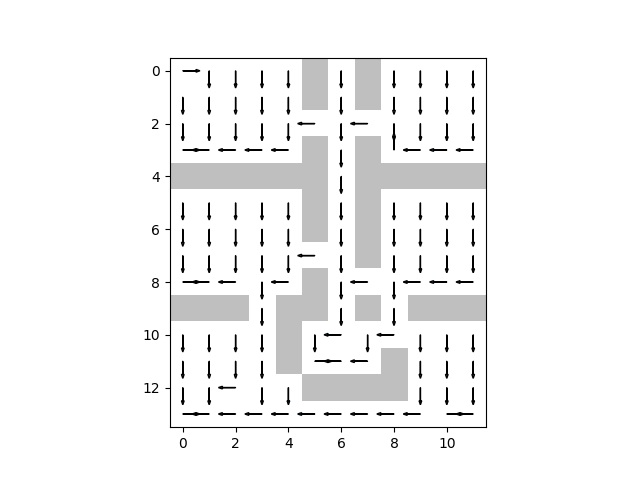}
    \end{subfigure}
    \end{figure}

    \begin{figure}[H]
    \centering
    \begin{subfigure}{.5\textwidth}
      \centering
      \includegraphics[width=\linewidth]{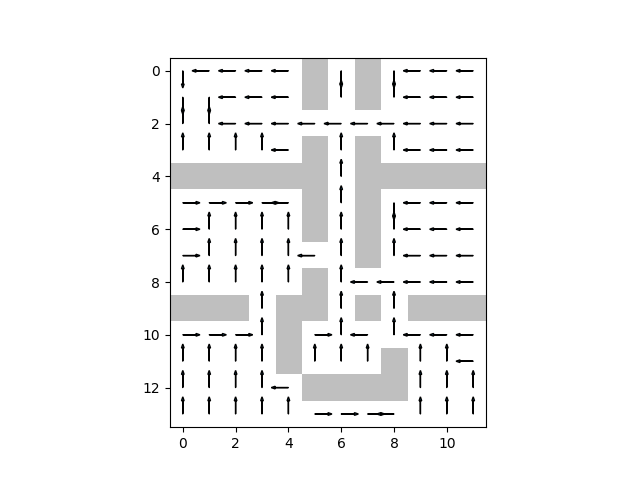}
    \end{subfigure}%
    \begin{subfigure}{.5\textwidth}
      \centering
      \includegraphics[width=\linewidth]{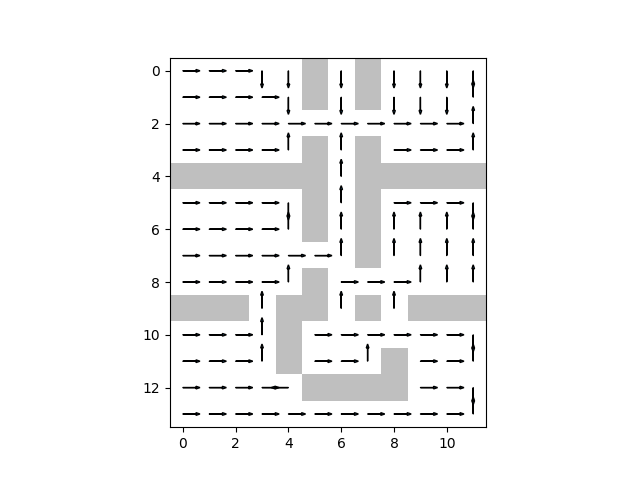}
    \end{subfigure}
    \end{figure}

    \begin{figure}[H]
    \centering
    \begin{subfigure}{.5\textwidth}
      \centering
      \includegraphics[width=\linewidth]{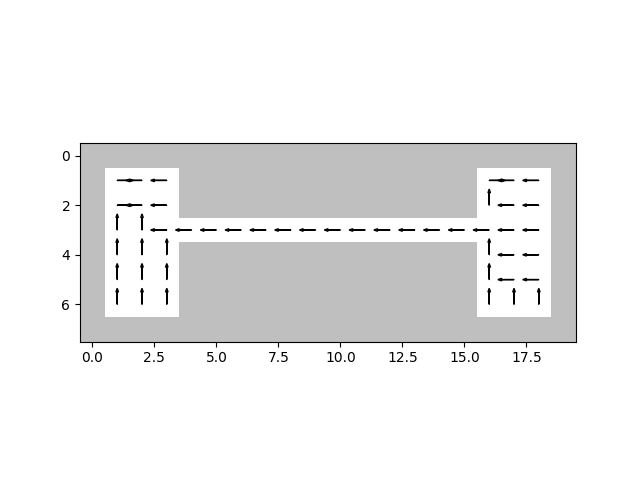}
    \end{subfigure}%
    \begin{subfigure}{.5\textwidth}
      \centering
      \includegraphics[width=\linewidth]{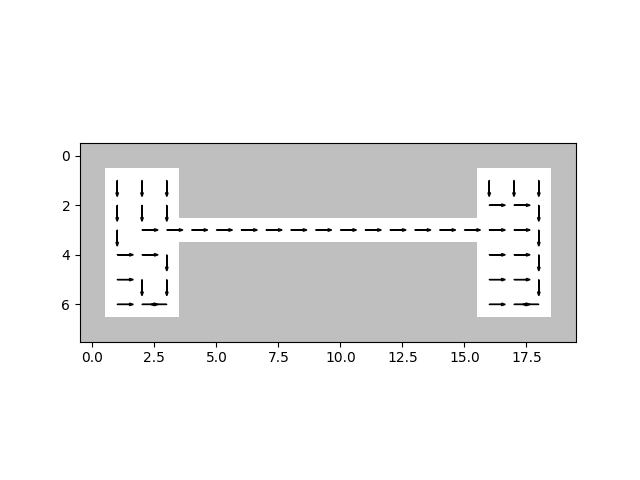}
    \end{subfigure}
    \end{figure}
    
    \newpage
    
    \section{Proposed Evaluation Metrics}
    
    \subsection{State-Visitation Count}
    State visitation counts give an estimate of the probability of visiting a particular state. Option policies with diverse state visitation counts can account for a multi-modal set of options. This metric can be used in continuous and high dimensional domains as well by using hash functions.
    
    \subsection{KL-Divergence}
    KL-Divergence between action distributions of agent and expert policy can be a good measure of how closely the agent is imitating expert.
    
    We also use KL-Divergence between two options policies as a measure of how different the options are. This can be used as regularizer to prevent mode collapse of options as described in Section 4.5.
    
    \subsection{Hinge Value Function Loss}
    We use hinge loss of agent's value function w.r.t that of an expert to measure how close to optimal is the behavior of agent w.r.t expert. Hinge loss measures the difference if the value function of the agent is lesser than that of expert policy at a state. Thus, it measures the maximal marginal difference.
\[L_{hinge}=\frac{1}{|\mathcal{S}|}\sum_{s\in\mathcal{S}}(\min (V(s),V^*(s)) - V^*(s))^2\]
    
    \subsection{Diffusion Time}
    This is the average over all pairs of states of the expected number of time steps required to go from one state to other using inferred options and primitive actions using a random walk.
    
    Small diffusion time implies an agent can cover a larger proportion of state space during initial exploration and also that options allow agent to get past bottleneck states to go to different ranges of state space.
    
    \subsection{T-SNE Embeddings}
    The representation of states can be projected into two-dimensional space using methods like T-SNE. The visualization of mapping between option and states over which it is accessible can be a good visual cue to determine the spatial variance of options.
    \begin{figure}[H]
        \centering
        \includegraphics[width=.6\textwidth]{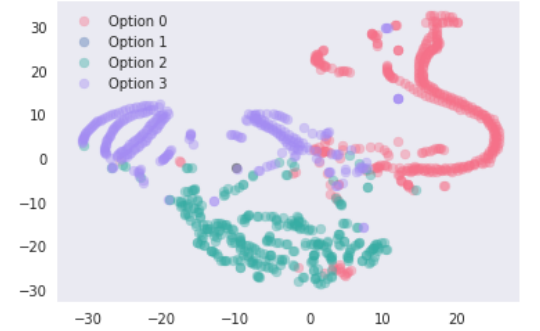}
        \caption{T-SNE Visualization}
    \end{figure}

\end{appendices}
\end{document}